# Extended Particle Swarm Optimization (EPSO) for Feature Selection of High Dimensional Biomedical Data

Ali Hakem Alsaeedi
College of Computer Science
and Information technology
University of Al-Qadisiyah
ali.alsaeedi@qu.edu.iq
Iraq

Adil L. Albukhnefis
College of Computer Science
and Information technology
University of Al-Qadisiyah
adil.lateef@qu.edu.iq
Iraq

Dhiah Al-Shammary
College of Computer Science
and Information technology
University of Al-Qadisiyah
d.alshammary@qu.edu.iq
Iraq

Muntasir Al-Asfoor
College of Computer Science
and Information technology
University of Al-Qadisiyah
muntasir.al-asfoor@qu.edu.iq
Iraq

**Abstract:**
This paper proposes a novel Extended Particle Swarm Optimization model (EPSO) that potentially enhances the search process of PSO for optimization problem. Evidently, gene expression profiles are significantly important measurement factor in molecular biology that is used in medical diagnosis of cancer types. The challenge to certain classification methodologies for gene expression profiles lies in the thousands of features recorded for each sample. A modified Wrapper feature selection model is applied with the aim of addressing the gene classification challenge by replacing its randomness approach with EPSO and PSO respectively. EPSO is initializing the random size of the population and dividing them into two groups in order to promote the exploration and reduce the probability of falling in stagnation. Experimentally, EPSO has required less processing time to select the optimal features (average of 62.14 sec) than PSO (average of 95.72 sec). Furthermore, EPSO accuracy has provided better classification results (start from 54% to 100%) than PSO (start from 52% to 96%).

# 1. Introduction

Particle Swarm Optimization Algorithm (PSO) is a well-known Evolutionary Algorithm (EA) that is highly successful in solving several hard optimization problems in different fields[1]. PSO suffers significantly like all other Evolutionary Algorithms from stagnation at local optimum as a result of lack in solution exploration and exploitation[2]. Exploration and exploitation are considered as core factors of EA[3]. They contribute together to move the search process towards the optimum solution. Both exploration and exploitation effect is based on jumping from local optimum to better local optimum and reduce the effect of stagnation on EA performance. In principle, improving solution exploration usually requires increasing diversity to detect new solutions[3], [4]. Moreover, improving the exploitation – The inheritance of useful features from previous iterations to reduce overall randomness – requires efficient features selection of predecessor iterations. On the other hand, features selection and reduction may generally enhance the performance of machine learning algorithms[1]. Metaheuristic is one of the popular methods for features selection. It is mainly based on randomness for searching optimal solutions. Metaheuristic has been recognized as fast, flexible, easy to implement, and successful in optimizing different fields[5]–[8].

## 1.1 Motivation

Recently, the number of data processing devices has increased tremendously that has led to a high volume of unwanted features sent to the data centers[9], [10]. Machine learning algorithms are significantly suffering from high dimensional data including many undesired features. Extracting and selecting relevant features can enhance data processing. Technically, reducing data dimension would optimize processing time, complexity, and memory management[2], [11].

PSO is an efficient technique to be applied in machine learning models for its simplicity, ease of development and potentials in selecting optimum features[1], [12]. The limitations of EA can be summarized in two aspects. First, EA often suffers from stagnation in advanced stages of the search process. This drawback is caused by the fact that PSO does not change its manner for the entire search process[2]. Second, it significantly falls into local optimum in high-dimensional data and has a low convergence rate in the iterative processes. In fact, our main motivation to develop a new potential model of PSO is inspired by the significant requirement of having a powerful optimizer of features selection for data mining and machine learning approaches. The limitations of PSO can be summarized in four main points:

- **High Stagnation Probability**: PSO is often suffering from stagnation during the search process. Therefore, It usually uses the same search methodology to deal with the search stages from the beginning to the advanced stages. Technically, PSO can lose diversity production in new candidate solutions in search process. This may cause weakness in exploration and lead the PSO into stagnation[13].
- **Stuck with Local optimum**: PSO can be significantly trapped into local optimum especially when facing a high-dimensional problem.
- **High Time Requirement**: PSO takes long time to find the optimal solution of given objective function. Although, PSO has a few numbers of variables, it still requires high processing time when treating complex problems.
- **Inpersistent Results**: Generally, PSO gives high variant results especially when dealing with a high dimensional objective function.

### 1.2 Contribution:

In this paper, a novel Extended Particle Swarm Optimization model (EPSO) is proposed which potentially enhances the search process of PSO for optimization search problem. We have assumed that the population pool is divided into two groups and applied different search methodologies on each group. The size of each group is inversely proportional to the size of the other one. The proposed method has provided several promising achievements:

- **Low Stagnation Probability**: EPSO divides a population into two groups in order to promote the exploration and reduce the probability of falling in stagnation.
- **Unstuck with Local optimum**: The proposed system performs a hybrid algorithm in order to change its manner gradually during the search process. EPSO increases the diversity of new solutions which are generated during the search process. Therefore, it operates efficiently when facing high-dimensional problems.

- **Low Time Requirement**: EPSO takes a short time to find the optimal solution of a given objective function comparing to PSO.
- **Persistent Results**: the EPSO gives approximately close results when dealing with different dimensional objective functions.

### 1.4 Paper Organization

The rest of this paper has been organized as follows: Section (2) presents the related works. More after, The Wrapper Model and Particle Swarm Optimization (PSO) are presented in sections (3) and (4) respectively. Furthermore, section (5) illustrates the proposed EPSO. The main description of the dataset, experiments and empirical results are shown in section (6). The conclusion of the work is presented in section (7).

## 2. Related works

Several studies have been proposed to develop feature selection techniques. Xiang yang *et. al*[1] used rough sets and binary version of PSO for feature selection. They note the exploration in PSO laying by enhancing the velocity of the particle. In addition, It leads the particle to the best possible target of the search process on a good solution. The velocity of new particles is considered when the value is less than the velocity of the best solution. Otherwise, it is replaced by the difference between the current and best solutions.

Chuang *et. al* [11]used the Binary version of Particle Swarm Optimization (BPSO) for feature selection and K-NN as an object function of classification problems. They enhanced classification progress by reducing the dimensions of data. The proposed method improves the prediction of K-NN without modifying the PSO.

Y. Zhang *et. al* [14] have applied the binary version of the stochastic optimizer (PSO) to rank optimal subset valid features. The modification of their method focused on increasing the diversity of PSO in the searching process. The mutation operation is used after updating the position of the particle. The value of the gene is flipped when the random threshold exceeds or equal to the probability of the mutation factor.

More after, Y. Zhang *et. al*[15] have introduced a new modification of PSO by increasing the exploration of the proposed algorithm. They modified the strategy of the algorithm by adding two characteristics. firstly, the mutation factor is added to update new solutions, when their value exceeds the random threshold [-1,1]. The probability of the applied mutation is increasing gradually during search progress. The second characteristic was added to improve the search process which basically based on the best solution and the difference between two random solutions which are picked from the solutions pool.

## 3. Wrapper Model

Wrapper model is a significant feature selection method[12]. It selects features based on trial and error, then updates corresponding features after each iteration. Furthermore, the features are selected randomly. The metaheuristic techniques could be suitable for ranking optimal features [12]. PSO is developed in the proposed system as a potential features selection for the wrapper model. Figure 1 illustrates the principles of the wrapper model for selecting the optimum features.

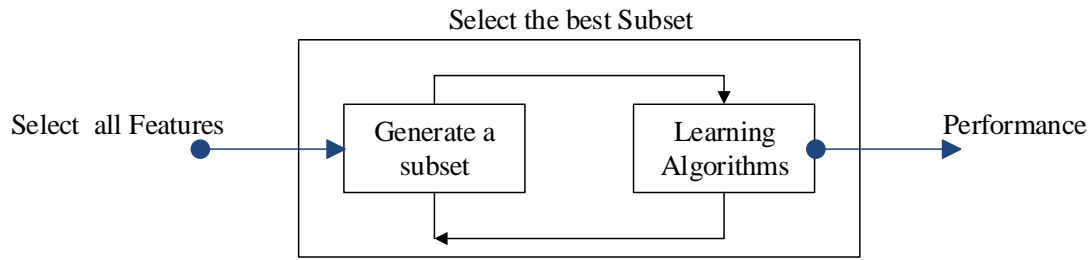

**Figure 1:** Wrapper Model

## 4. Particle Swarm Optimization (PSO)

PSO is a stochastic Evolutionary Algorithm (EA) that simulating animals social behavior which are living in groups[16], [25]. PSO has been preferred over other EA as a result of its simple mathematical model development and using of few variables. It structurally resembles evolutionary optimizers by starting randomly and partake of all participants to find the optimal solution[2]. The search engine of PSO is substantially based on a population of particles. Each particle starts with a random position and zero velocity. Particle velocity and position are updated dynamically during search progress. Each PSO's particle stores two values (current solution and best solution that figure out by itself). The best solution of PSO's particle is called the local best optimum ($p_{best}$). The global best solution ($g_{best}$) is the best solution among $p_{best}$s and is updated after a complete optimization iteration. Equation 1 calculates the velocity of particles [8].

$$V_i^d(t+1) = w(t)V_i^d(t) + c_1 r_1 \left(pbest_i^d - x_i^d(t)\right) + c_2 r_2 (gbest^d - x_i^d(t)) \quad (1)$$

where: $r1$ and $r2$ are random variables in the range [0, 1]. $c1$ and $c2$ are positive coefficients. $w$ is the inertia weight. $v_i^d(t), x_i^d(t)$ indicates the velocity and position of $i^{th}$ particle at iteration t in $d^{th}$ dimension respectively.

Equation 2 is applied to find the new value of particle position (candidate solution) [8].

$$x_i^{t+1} = x_i^t + v_i^{t+1} \quad (2)$$

Where: $x_i^t$ is old particle position and $x_i^{t+1}$ is a new particle position.

## 5. Proposed Extended Particle Swarm Optimization (EPSO)

EPSO is inspired by the principle of life restoration. Technically, the life of a swarm search starts directly after the end of the previous swarm search. It is developed to distribute the population into two groups. Each group uses a different search methodology. First search group is based on the traditional PSO methodology. On the other hand, a new search methodology that is developed based on the arithmetic crossover for the second search group. In fact, multiple search methodologies are proposed with the aim to promote diversity in the next generations. Therefore, the exploration of a new solution has increased by high diversity[3], [4], [17]. Moreover, EPSO like other EAs consists of four stages: initialization, population distribution, updating population, and termination. Figure 2 shows the main components of the proposed EPSO.

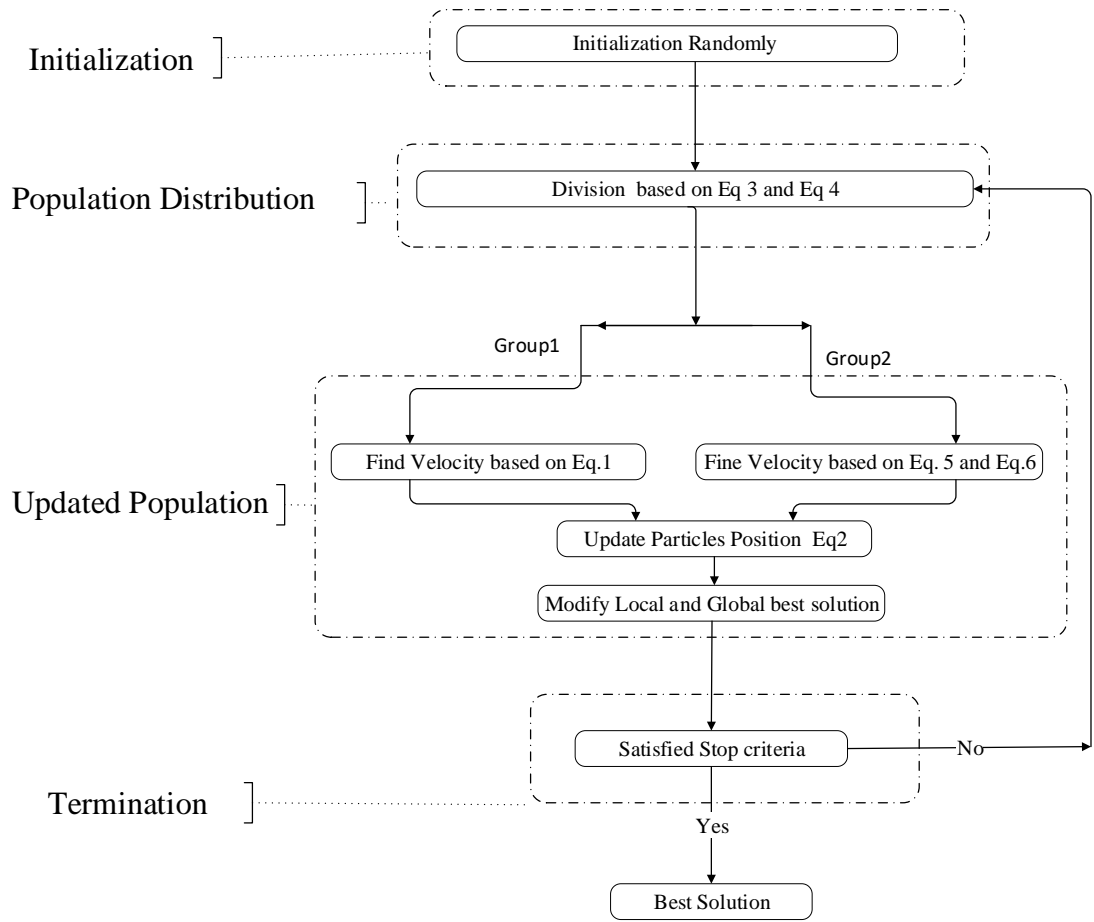

**Figure 2:** Main component of proposed EPSO

### 5.1 Initialization

The initialization of the proposed algorithm is mainly based on setting the size of the random population and variant percentage size of each population group. A random population starts with a size set by the user and a specific dimension according to the optimization problem. Furthermore, the user sets maximum and minimum percentage size of extending (second) PSO population group, and maximum and minimum number of genes for mutation of extending PSO group. The rest of the population represent the size of the first group.

### 5.2 Population Distribution

In order to reduce the probability of stagnation in the search process, the proposed algorithm distributes the population into two groups ($EPSO_{G1}$, $EPSO_{G2}$) that have inverse

relation in terms of size. Equations 3 and 4 [18] calculates the size of groups after each search iteration :

$$EPSO_{G1} = round\left(\left(g_{pini} - \left(\frac{current\ iteration}{Max_{iteration}}\right)^2 \times (g_{pini} - g_{pfini})\right) * popsize\right) \quad (3)$$

$$EPSO_{G2} = population\ size - EPSO_{G1} \quad (4)$$

Where : $g_{pini}$ , $g_{pfine}$ are a maximum and minimum percentages of $EPSO_{G1}$ respectively

Figure3 illustrates how the population is distributed during the search progress

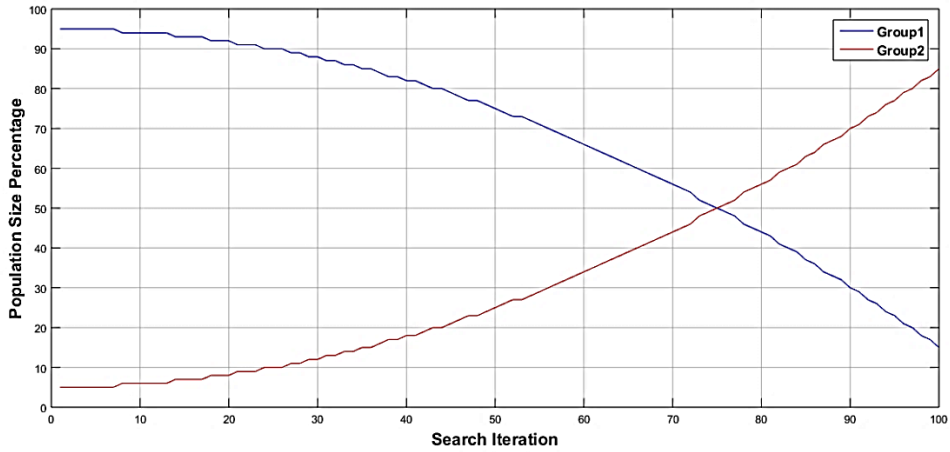

**Figure 3:** Population Distribution

### 5.3 Particles Searching Methodology

The velocity of a particle represents the mutation value used to move the particle by adding its value to particle position computing a new position. Furthermore, new particles velocity and position in $EPSO_{G1}$ are computed by Equations 5 and 6 respectively. On the other hand, particles velocity in $EPSO_{G2}$ is modified by selecting random $m$ genes ($EPSO_{G2,Genes}$) from corresponding particle. Equation 5 [19] illustrates the computing of new particle genes. Moreover, velocity value is proportional to the size of $EPSO_{G2}$.

$$EPSO_{G2,Genes} = round(M_{min} + \left(\frac{current\ iteration}{Max_{iteration}}\right)^2 * (M_{max} - M_{min})) \quad (5)$$

Where: $M_{max}$ , $M_{min}$ are maximum and minimum genes of $EPSO_{G2}$ particles respectively.

Figure 4 illustrates the relation between number of mutation gens in $2^{nd}$ group ($EPSO_{G2}$) and search progress.

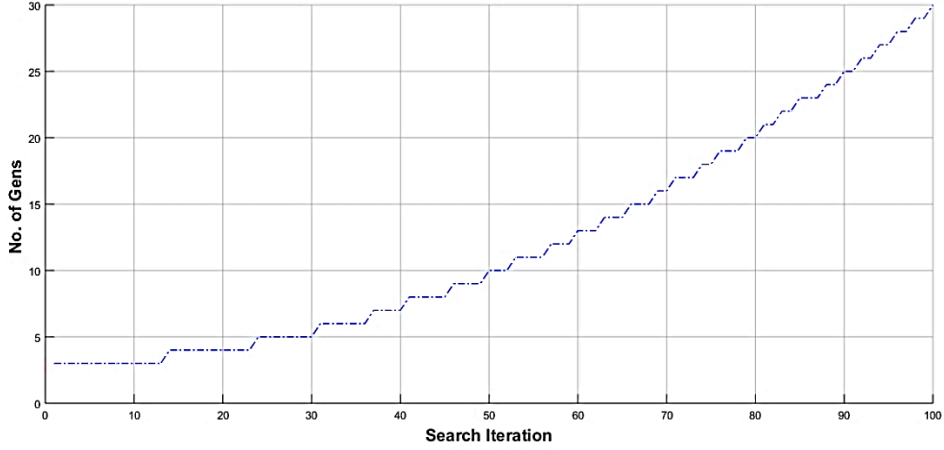

**Figure 4:** Number Mutation gens in $2^{nd}$ group ($EPSO_{G2}$)

Final new particle velocity ($V_i^d(t+1)$) of $EPSO_{G2,Genes}$ is computed by Eq.6:

$$V_i^d(t+1) = \alpha\, gbest + (1 - \beta\, V_i^d(t)_i)\, pbest_i \quad , \quad i \in EPSO_{G2,Genes} \tag{6}$$

Where: $\alpha, \beta$ random value [-1,1]

### 5.4 Features Selection Methodology:

Wrapper model is based on binary metaheuristic algorithms. Therefore, searching the space is limited by the binary interval [0,1]. The proposed EPSO has improved the exploration by expanding space searching interval used to check the corresponding subset features. The applied space searching interval range is [-1,+1]. Finally, the threshold for feature selection is determined by the user for each experiment[20]. For example, the threshold value of 0.5 means omitting all values less than the applied threshold (0.5) and take positions in a current-generation for values greater than the same threshold. The proposed system terminates when searching process iteration exceeded the limitation of the maximum iteration.

# 6 Discussion and Experimental Results

The evaluation strategy is based on testing the proposed model using the functions benchmark Congress on Evolutionary Computation 2015 (CEC'15) [21]. Furthermore, the proposed model is applied on a high dimensional medical dataset developed and proposed by[23], [24]. Finally, the proposed EPSO is compared with traditional PSO and pure classification. The results have shown higher accuracy in comparison with other models.

## 6.1 Congress on Evolutionary Computation Benchmark

Congress on Evolutionary Computation Benchmark is applied to EPSO and traditional PSO in order to investigate our model capabilities in finding the optimum solution.

### 6.1.1 Benchmark Description

The CEC'15 is the potential benchmark to evaluate Evolutionary Algorithms. It has fifteen functions divided into four groups (Unimodal Functions Group, Simple Multimodal Functions, Hybrid Functions, and Composition Functions). Figure 5 shows the groups of CEC'15

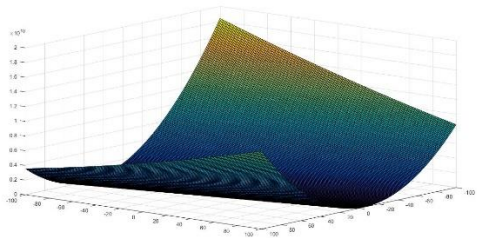
F1: Rotated High Conditioned Elliptic Function

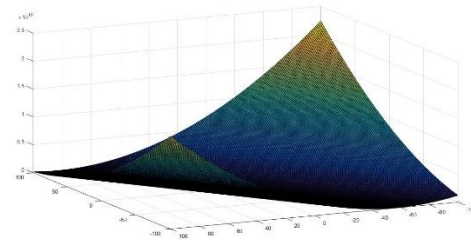
F2Rotated Cigar Function:

**A-Unimodal Functions**

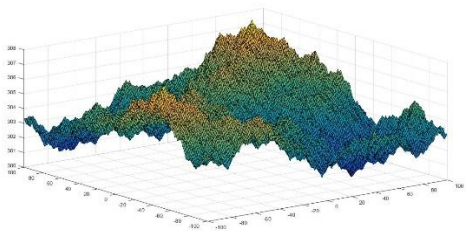
F3: Shifted and Rotated Ackley's Function

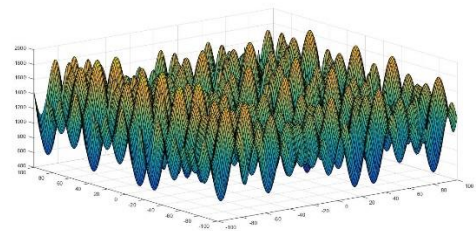
F4: Shifted and Rotated Rastrigin's Function

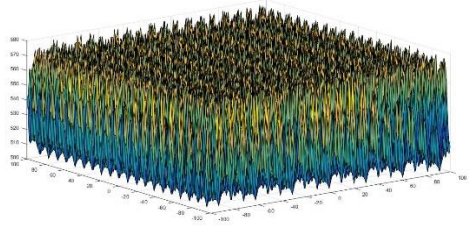

F5: Shifted and Rotated Schwefel's Function

**B-Simple Multimodal Functions**

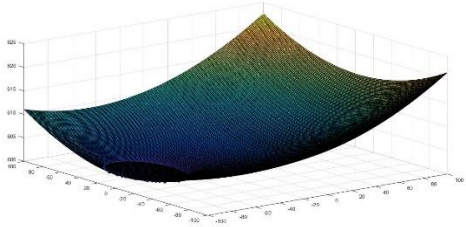

F6: Hybrid Function 1 (N=3)

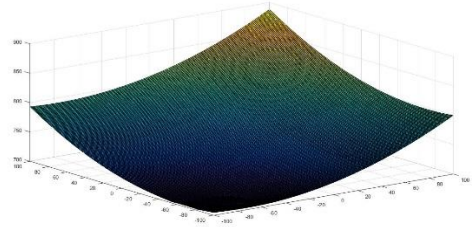

F7: Hybrid Function 2 (N=4)

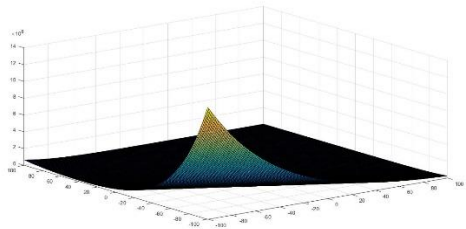

F8: Hybrid Function 3 (N=5)

**C: Hybrid Functions**

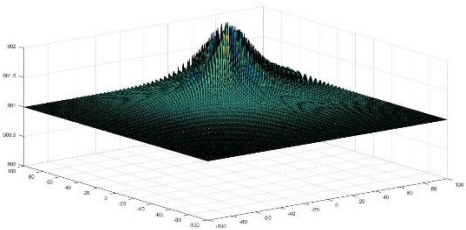

F9: Composition Function 1 (N=3)

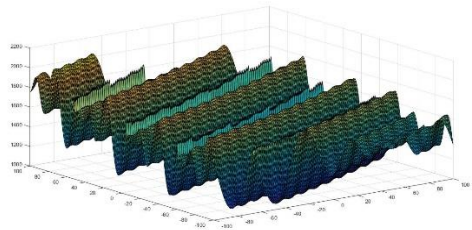

F10: Composition Function 2 (N=3)

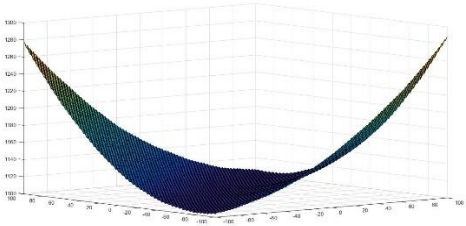

F11: Composition Function 3 (N=5)

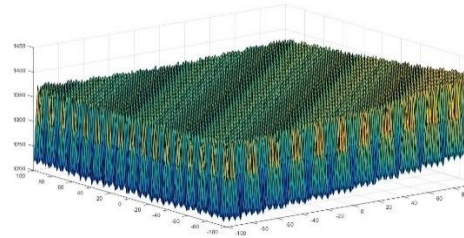

F12: Composition Function 4 (N=5)

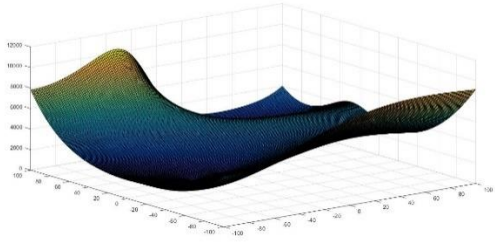
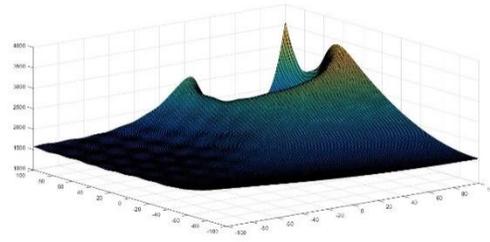

F13: Composition Function 5 (N=5)   F14: Composition Function 6 (N=7)

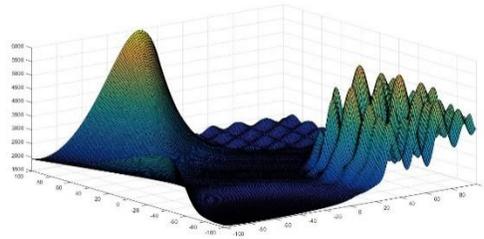

F15: 15 Composition Function 7 (N=10)
**D:Composition Functions**

**Figure 5:** CEC'15 Benchmark Function Groups [21]

### 6.1.2 Results

Both EPSO and traditional PSO have been executed thirty times with the aim to investigate model persistency in searching for the optimum solution. Five evaluation criterion: best, worst, mean, median, and standard deviation are calculated for the outcome results. The evaluation analysis is based on the capability to find the minimum value of best, worst, mean, and median. In fact, the highest standard deviation is observed as the best performance. The standard deviation is calculated by finding the difference between the first and final stage of search progress. Technically, higher standard deviation indicates better exploration[3], [19]. Table 1 shows the results for CEC'15 functions over 30 independent runs by EPSO and PSO.

**Table 1:** Comparative statistics of optimization results (EPSO and PSO) of CEC'15 functions over 30 independent runs

| Function No. | EPSO | | | | | PSO | | | | |
|---|---|---|---|---|---|---|---|---|---|---|
| | Mean | Median | Std | Best | Worst | Mean | Median | Std | Best | Worst |
| F1 | 351.28 | 350.7672 | 1.04095 | 350.7459 | 357.9543 | 351.28 | 350.7772 | 1.04095 | 350.7459 | 357.9543 |
| F2 | **253.136** | **250.3717** | 6.212173 | **250.3515** | **297.8943** | 253.249 | 250.8777 | 5.514791 | 250.587 | 302.1339 |
| F3 | **330.384** | **330.3032** | 0.147118 | **330.285** | **331.3303** | 330.399 | 330.3428 | 0.132978 | 330.3106 | 331.3317 |
| F4 | **531.254** | **529.0699** | 4.902492 | **529.0699** | **563.227** | 534.343 | 531.9586 | 4.43649 | 529.0699 | 563.227 |
| F5 | **504.22** | **503.9214** | 0.6592 | **503.8291** | **509.963** | 504.258 | 504.0068 | 0.676596 | 503.8291 | 509.963 |
| F6 | **606.504** | **606.5015** | 0.011324 | **606.5014** | **606.573** | 606.506 | 606.5021 | 0.009582 | 606.5019 | 606.5742 |
| F7 | 849.873 | 849.7148 | 0.38846 | 849.7065 | 852.0364 | **849.868** | **849.7065** | 0.32162 | **849.7065** | **852.2176** |
| F8 | 862.202 | 857.47 | 11.17502 | 857.1243 | 942.1431 | **861.711** | **857.896** | 9.376672 | **857.6942** | **937.5275** |
| F9 | **909.243** | **909.1889** | 0.110818 | **909.1798** | **910.0147** | 909.254 | 909.2149 | 0.096148 | 909.198 | 910.0149 |
| F10 | **1330.73** | **1317.791** | 28.39088 | **1317.694** | **1526.316** | 1344.13 | 1333.878 | 26.24366 | 1332.905 | 1551.636 |
| F11 | **1210.82** | **1207.834** | 6.132492 | **1207.834** | **1252.941** | 1210.85 | 1208.241 | 5.789778 | 1207.855 | 1253.729 |
| F12 | **1213.08** | **1203.118** | 22.70002 | **1203.67** | **1379.215** | 1214.24 | 1204.066 | 22.14039 | 1202.938 | 1379.326 |
| F13 | 1303 | 1301.685 | 2.742172 | 1301.593 | 1327.5 | **1302.27** | **1301.23** | 2.517544 | **1301.112** | **1326.97** |
| F14 | **1438.76** | **1437.084** | 3.461823 | **1437.02** | **1462.678** | 1440.47 | 1439.107 | 3.100264 | 1439.02 | 1463.594 |
| F15 | **1546.59** | **1543.784** | 5.543503 | **1543.634** | **1582.805** | 1547.16 | 1545.269 | 4.880155 | 1544.37 | 1585.268 |

In the first group Unimodal (consist of two functions), EPSO has succeeded to find the best result in the second function and same achievement of traditional PSO for first function. Simple Multimodal Functions have several local optimums. Therefore, high exploration is necessary to avoid the stagnation phenomena at a local optimum. The proposed algorithm has outperformed traditional PSO with all functions of Multimodal group. Furthermore, in function 6 of Hybrid functions, EPSO has clearly achieved higher performance than PSO. On the other hand, in functions 7 and 8 PSO has slightly achieved better performance than EPSO. However, PSO has lower standard deviation. In the last group, there are several local optimums, therefore, the high exploration is required**.** The proposed EPSO has higher exploration and achieved better results in functions 9, 10, 11, 12, 14, and 15, while the PSO produced the best result in function 13 only. The PSO falls into local optimum because it has a low convergence rate in the iterative process.

To sum up, the proposed EPSO has achieved potential exploration evident by the obtained higher standard deviation. These significant results can be explained by the EPSO capability to generate new best solutions and jump from local optimum to better local optimum. Finally, based on the overall evaluation criterion, EPSO has succeeded in most CEC'15 benchmark functions and can be an alternative to the current optimization methods. Figures 6 and 7 illustrate the possibility of EPSO to overcome the stagnation experienced by the PSO algorithm.

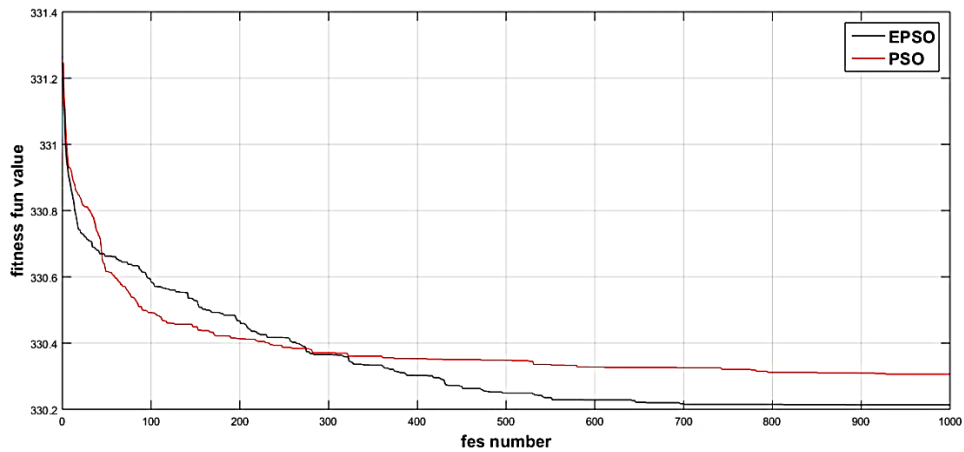

Figure 6: Comparison search performance of EPSO and PSO on Function 3

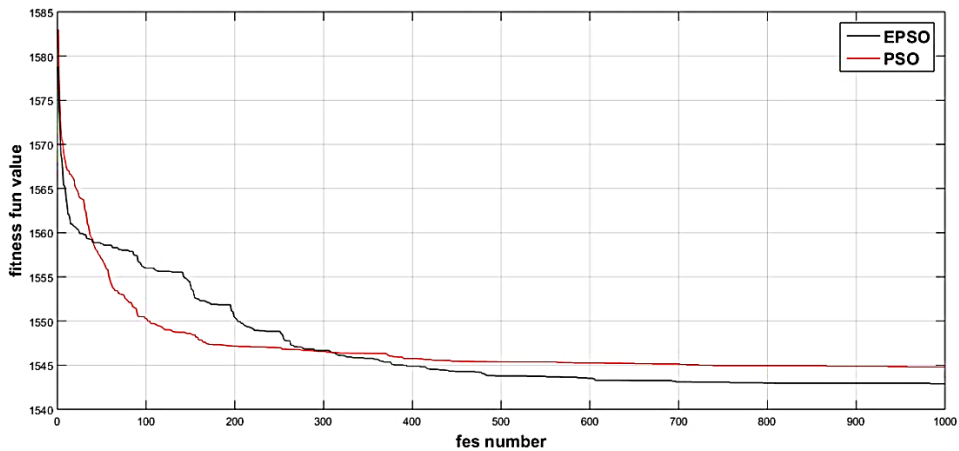

Figure 7: Comparison of search performance of EPSO and PSO on Function 15

## 6.2 Biomedical Dataset Classification

With the aim of investigate our proposed model performance in real-life problems, we have tested EPSO as features selection for medical information. At the same time, the feature selection based on traditional PSO and without feature selection are applied in order to compare with EPSO.

### 6.2.1 Dataset Description

The dataset which has been used in this work consists of twenty medical groups with thousands of features. They are collected from two websites[22]: first one is at [23] that has eleven groups (14_tumors, 11_tumors, 9_tumors, brain_tumor1, brain_tumor2, leukemia1, leukemia2, lung_cancer, SRBCT, prostate_tumor, and DLBCL), and second dataset has other nine groups from website [24]

Technically, dataset complexity is usually represented by the parameter $C.F/O$ as C is the number of classes, F is the number of features, and O is the number of observations. Generally, a bigger number of features than the number of observations means high dataset complexity. Table 2 illustrates the dataset details with complexity measurements.

**Table 2:** Dataset details [22]

| Sq | Dataset Name | No. Class | No. Samples | No. Features | C.F/O |
|---|---|---|---|---|---|
| 1 | 14_tumors | 26 | 308 | 15,010 | 1267 |
| 2 | 9_tumors | 9 | 60 | 5726 | 859 |
| 3 | Brain_tumor2 | 4 | 50 | 10,367 | 829 |
| 4 | 11_tumors | 11 | 174 | 12,533 | 792 |
| 5 | Nci | 8 | 61 | 5244 | 688 |
| 6 | Brain_tumor2 | 5 | 42 | 5597 | 666 |
| 7 | Leukemia2 | 3 | 72 | 11,225 | 468 |
| 8 | Brain_tumor1 | 5 | 90 | 5920 | 329 |
| 9 | Lung_cancer | 4 | 203 | 12,600 | 310 |
| 10 | Adenocarcinoma | 2 | 76 | 9868 | 260 |
| 11 | Leukemia1 | 3 | 72 | 5327 | 222 |
| 12 | Prostate_tumor | 2 | 102 | 10,510 | 206 |
| 13 | Lymphoma | 3 | 62 | 4026 | 195 |
| 14 | Leukemia | 2 | 38 | 3051 | 161 |
| 15 | Breast3 | 3 | 95 | 4869 | 154 |

Table 2 (cont'd)

| | | | | | |
|---|---|---|---|---|---|
| 16 | DLBCL | 2 | 77 | 5469 | 142 |
| 17 | Breast2 | 2 | 77 | 4869 | 126 |
| 18 | Prostate | 2 | 102 | 6033 | 118 |
| 19 | SRBCT | 4 | 82 | 2308 | 113 |
| 20 | Colon | 2 | 62 | 2000 | 65 |

### 6.2.2 Results

Generally, the few numbers of features selection does not necessarily refer to a good characteristic of the features selection algorithm. The main principle comparison is investigating the number of valid features and their effect on the accuracy of the objective function. KNN is applied as an objective function with the aim to test/evaluate and compare the performance of PSO and EPSO. The number of neighbors (K) is set by one in order to reduce the objective function noise.

Table 3. shows the performance metrics applied to the proposed algorithm and PSO. This table compares the resultant time and accuracy of 1NN (all features engagement), PSO and EPSO. Clearly, EPSO has achieved higher classification accuracy rates than others. Moreover, this fact is highly confirmed when the number of features exceeded the number of observations (the high value of C.F/O). In other words, the proposed EPSO feature selection has consistently required fewer features to deliver accurate classification results in high-dimensional datasets and demonstrated its value as a potential alternative to PSO.

**Table 3:** Classification performance with full feature and feature selection by PSO and EPSO (standard deviations, accuracy, and required time)

| Sq | Dataset Name | C.F/O | 1NN | PSO | | | | EPSO | | | |
|---|---|---|---|---|---|---|---|---|---|---|---|
| | | | | Features | Accuracy | Std | Time (sec) | Features | Accuracy | Std | Time (sec) |
| 1 | 14_tumors | 1267 | 0.57 | 202 | 0.72 | 0.08 | 328.91 | 83 | **0.75** | 0.05 | 187.83 |
| 2 | 9_tumors | 859 | 0.51 | 86 | 0.5 | 0.35 | 120.78 | 35 | **0.63** | 0.23 | 78.64 |
| 3 | Brain_tumor2 | 829 | 0.86 | 75 | 0.68 | 0.7 | 183.20 | 23 | **0.74** | 0.23 | 80.92 |
| 4 | 11_tumors | 792 | 0.82 | 131 | 0.79 | 0.16 | 128.63 | 99 | **0.87** | 0.07 | 98.22 |
| 5 | Nci | 688 | 0.72 | 145 | 0.68 | 0.45 | 135.22 | 38 | **0.74** | 0.31 | 65.21 |
| 6 | Brain_tumor2 | 666 | 0.71 | 100 | 0.73 | 0.15 | 98.47 | 60 | **0.78** | 0.06 | 59.08 |
| 7 | Leukemia2 | 468 | 0.92 | 85 | 0.96 | 0.9 | 112.88 | 22 | **0.98** | 0.1 | 43.78 |

Table 3 (cont'd)

| 8 | Brain_tumor1 | 329 | 0.86 | 58 | 0.85 | 0.19 | 64.85 | 30 | **0.9** | 0.11 | 40.83 |
|---|---|---|---|---|---|---|---|---|---|---|---|
| 9 | Lung_cancer | 310 | 0.9 | 10 | 0.91 | 0.13 | 50.12 | 25 | **0.94** | 0.1 | 61.22 |
| 10 | Adenocarcinoma | 260 | 0.81 | 53 | 0.77 | 0.06 | 87.34 | 14 | **0.96** | 0.04 | 25.09 |
| 11 | Leukemia1 | 222 | 0.89 | 37 | 0.82 | 0.07 | 53.47 | 25 | **0.94** | 0.07 | 44.57 |
| 12 | Prostate_tumor | 206 | 0.77 | 46 | 0.85 | 10 | 78.32 | 25 | **0.95** | 0.09 | 36.89 |
| 13 | Lymphoma | 195 | 0.98 | 81 | 0.95 | 0 | 93.12 | 73 | **1** | 0.04 | 91.22 |
| 14 | Leukemia | 161 | 0.89 | 8 | 0.9 | 0.07 | 23.02 | 17 | **0.98** | 0.01 | 42.89 |
| 15 | Breast3 | 154 | 0.55 | 11 | **0.62** | 0.21 | 46.16 | 20 | 0.57 | 0.16 | 51.28 |
| 16 | DLBCL | 142 | 0.86 | 32 | 0.89 | 0.13 | 63.55 | 44 | **0.97** | 0.05 | 66.21 |
| 17 | Breast2 | 126 | 0.62 | 28 | 0.52 | 0.21 | 37.69 | 15 | **0.54** | 0.11 | 28.55 |
| 18 | Prostate | 118 | 0.83 | 5 | **0.93** | 0.09 | 17.25 | 9 | 0.9 | 0.12 | 16.22 |
| 19 | SRBCT | 113 | 0.92 | 40 | **0.98** | 0.02 | 75.23 | 35 | 0.96 | 0.01 | 62.48 |
| 20 | Colon | 65 | 0.73 | 18 | **0.87** | 0.13 | 49.03 | 18 | 0.87 | 0.1 | 40.99 |

## 7. Conclusion

In conclusion, EAs are usually suffering from stagnation during search progress as a result of losing the initiative to enhance its exploration. Therefore, increasing the randomness in advanced stages of the search process is necessary. In this paper, we have developed a new version of PSO that has reduced the stagnation status significantly. Our proposed Extended PSO (EPSO) has achieved better results through distributing the population into two groups with simultaneous search processes. Each group with different size inverse to the other group size. Moreover, the first group is applied with regular PSO search while the other group is applied with highly random search. The proposed model is compared with traditional PSO and full features classification. Experimentally, the proposed EPSO has outperformed other models and proven to be a potential optimization alternative. For future work, a proactive random model would be applied in order to adapt the variable search volume in most optimization techniques.

# References

[1] X. Wang, J. Yang, X. Teng, W. Xia, and R. Jensen, "Feature selection based on rough sets and particle swarm optimization," Pattern Recognit. Lett., vol. 28, no. 4, pp. 459–471, 2007.

[2] A. H. Jabor and A. H. Ali, "Dual Heuristic Feature Selection Based on Genetic Algorithm and Binary Particle Swarm Optimization," J. Univ. BABYLON pure Appl. Sci., vol. 27, no. 1, pp. 171–183, 2019.

[3] M. Lones, Sean Luke: essentials of metaheuristics, vol. 12, no. 3. 2011.

[4] X. S. Yang, "Bat algorithm: Literature review and applications," Int. J. Bio-Inspired Comput., vol. 5, no. 3, pp. 141–149, 2013.

[5] S. Ibrahim, N. A. Wahab, F. S. Ismail, and Y. M. Sam, "Optimization of artificial neural network topology for membrane bioreactor filtration using response surface methodology," IAES Int. J. Artif. Intell., vol. 9, no. 1, pp. 117–125, 2020.

[6] F. Lü, C. Qin, and Yunpeng, "Particle swarm optimization-based BP neural network for UHV DC insulator pollution forecasting," J. Eng. Sci. Technol. Rev., vol. 7, no. 1, pp. 132–136, 2014.

[7] N. F. Fadzail, S. M. Zali, M. A. Khairudin, and N. H. Hanafi, "Stator winding fault detection of induction generator based wind turbine using ANN," vol. 19, no. 1, pp. 126–133, 2020.

[8] S. Ding, H. Li, C. Su, J. Yu, and F. Jin, "Evolutionary artificial neural networks: A review," Artif. Intell. Rev., vol. 39, no. 3, pp. 251–260, 2013.

[9] J. P. D. Comput, D. Al-shammary, I. Khalil, Z. Tari, and A. Y. Zomaya, "Fractal self-similarity measurements based clustering technique for SOAP Web messages," J. Parallel Distrib. Comput., vol. 73, no. 5, pp. 664–676, 2013.

[10] R. Kaubruegger, L. Pastori, and J. C. Budich, "Chiral topological phases from artificial neural networks," Phys. Rev. B, vol. 97, no. 19, pp. 1–8, 2018.

[11] L. Y. Chuang, H. W. Chang, C. J. Tu, and C. H. Yang, "Improved binary PSO for feature selection using gene expression data," Comput. Biol. Chem., vol. 32, no. 1, pp. 29–38, 2008.

[12] N. Sánchez-Maroño, A. Alonso-Betanzos, and M. Tombilla-Sanromán, "Filter methods for feature selection - A comparative study," Lect. Notes Comput. Sci.

(including Subser. Lect. Notes Artif. Intell. Lect. Notes Bioinformatics), vol. 4881 LNCS, pp. 178–187, 2007.

[13] S. S. Aote, M. M. Raghuwanshi Principal, and R. Latesh Malik HOD, "A Brief Review on Particle Swarm Optimization: Limitations & Future Directions," Int. J. Comput. Sci. Eng., vol. 2, no. 05, pp. 2319–7323, 2013.

[14] Y. Zhang, S. Wang, P. Phillips, and G. Ji, "Binary PSO with mutation operator for feature selection using decision tree applied to spam detection," Knowledge-Based Syst., vol. 64, pp. 22–31, 2014.

[15] Y. Zhang, D. W. Gong, X. Y. Sun, and Y. N. Guo, "A PSO-based multi-objective multi-label feature selection method in classification," Sci. Rep., vol. 7, no. 1, pp. 1–12, 2017.

[16] R. Eberhart and J. Kennedy, "New optimizer using particle swarm theory," Proc. Int. Symp. Micro Mach. Hum. Sci., pp. 39–43, 1995.

[17] L. Wang, X. Fu, M. I. Menhas, and M. Fei, "A modified binary differential evolution algorithm," Lect. Notes Comput. Sci. (including Subser. Lect. Notes Artif. Intell. Lect. Notes Bioinformatics), vol. 6329 LNCS, no. PART 2, pp. 49–57, 2010.

[18] J. L. Rueda and I. Erlich, "Evaluation of the mean-variance mapping optimization for solving multimodal problems," Proc. 2013 IEEE Symp. Swarm Intell. SIS 2013 - 2013 IEEE Symp. Ser. Comput. Intell. SSCI 2013, pp. 7–14, 2013.

[19] A. H. Al-saeedi, "Binary Mean-Variance Mapping Optimization Algorithm (BMVMO)," J. Appl. Phys. Sci., vol. 2, no. 2, pp. 42–47, 2016.

[20] R. N. Khushaba, A. Al-Ani, and A. Al-Jumaily, "Differential Evolution based feature subset selection," Proc. - Int. Conf. Pattern Recognit., 2008.

[21] G. Wu, R. Mallipeddi, and P. N. Suganthan, "Problem Definitions and Evaluation Criteria for the CEC 2017 Competition on Constrained Real- Parameter Optimization," Tech. Report, Nanyang Technol. Univ., no. November 2014, pp. 1–16, 2016.

[22] C. H. Park and S. B. Kim, "Sequential random k-nearest neighbor feature selection for high-dimensional data," Expert Syst. Appl., vol. 42, no. 5, pp. 2336–2342, 2015.

[23] www.gems-system.org

[24] www.ligarto.org/rdiaz/ Papers/rfVS/randomForestVarSel.html .

[25]  University of Ioannina, "Research Interests", accessed February 14 2020, www.cs.uoi.gr